\documentclass[final,3p,times, 12pt]{elsarticle}
\usepackage[hidelinks]{hyperref}

\usepackage{amssymb}
\usepackage{amsmath}
\usepackage{booktabs}
\usepackage{multirow}

\begin{document}

\begin{frontmatter}



\title{QuanvNeXt: An end-to-end quanvolutional neural network for EEG-based detection of major depressive disorder}

\author[inst1]{Nabil Anan Orka}
\affiliation[inst1]{organization={The University of Queensland},
            city={St Lucia},
            postcode={4072}, 
            state={QLD},
            country={Australia}}

\author[inst2]{Ehtashamul Haque}
\affiliation[inst2]{organization={BRAC University},
            city={Dhaka},
            postcode={1212}, 
            state={Dhaka Division},
            country={Bangladesh}}

\author[inst3]{Maftahul Jannat}
\affiliation[inst3]{organization={Bangladesh Medical University},
            city={Dhaka},
            postcode={1000}, 
            state={Dhaka Division},
            country={Bangladesh}}

\author[inst4]{Md Abdul Awal}
\affiliation[inst4]{organization={University of Southern Queensland},
            city={Toowoomba},
            postcode={4350}, 
            state={QLD},
            country={Australia}}


\author[inst1,inst5]{Mohammad Ali Moni}
\affiliation[inst5]{organization={Charles Sturt University},
            city={Orange},
            postcode={2795}, 
            state={NSW},
            country={Australia}}




\begin{abstract}
This study presents QuanvNeXt, an end-to-end fully quanvolutional model for EEG-based depression diagnosis. QuanvNeXt incorporates a novel Cross Residual block, which reduces feature homogeneity and strengthens cross-feature relationships while retaining parameter efficiency. We evaluated QuanvNeXt on two open-source datasets, where it achieved an average accuracy of 93.1\% and an average AUC-ROC of 97.2\%, outperforming state-of-the-art baselines such as InceptionTime (91.7\% accuracy, 95.9\% AUC-ROC). An uncertainty analysis across Gaussian noise levels demonstrated well-calibrated predictions, with ECE scores remaining low (0.0436, Dataset 1) to moderate (0.1159, Dataset 2) even at the highest perturbation ($\epsilon = 0.1$). Additionally, a post-hoc explainable AI analysis confirmed that QuanvNeXt effectively identifies and learns spectrotemporal patterns that distinguish between healthy controls and major depressive disorder. Overall, QuanvNeXt establishes an efficient and reliable approach for EEG-based depression diagnosis.
\end{abstract}




\end{frontmatter}



\section{Introduction}
With the world rapidly advancing toward quantum computing, researchers are adapting existing machine learning (ML) algorithms to accommodate quantum processes, a field broadly referred to as quantum machine learning (QML). In this study, we explore QML in a relatively unexplored mental health field, major depressive disorder (MDD).

MDD, more commonly known as clinical depression, is a disease that affects around 280 million people worldwide \cite{who_depression}. The health concerns of depression and its strong association with suicide have made MDD a growing avenue for studies \cite{who_suicide, Thaipisuttikul2014, Steffen2020}. Currently, research on depression is focused on developing an objective way to diagnose the mental illness. While the questionnaire-based approach to depression diagnosis focusing on the \cite{dsm5} or ICD-11 \cite{icd11} criteria is well-studied and has proven quite accurate, the subjectivity of the tests remains a concern. To develop objective diagnostic methods, data from two brain activity detection techniques, EEG and fMRI, are primarily analyzed. Integrating ML models into this process has further improved diagnostic accuracy across both modalities \cite{deaguir2019, pilmeyer2022}.

While fMRI provides higher-resolution brain activity scans, EEG-based depression diagnosis is gaining popularity due to its lower cost, portability, and ease of use \cite{yasin2021eeg}. So far, various ML models have been applied to different datasets, resulting in promising but varying accuracies. Early studies in this field focused on manually extracting features from EEG signals and training shallow ML models such as logistic regression, decision trees, support vector machines, and other ensemble methods \cite{hosseinifard2013, Mohammadi2015, Li2016}. As ML advanced, EEG-based depression classifiers evolved into more end-to-end systems using deep learning models like CNNs and RNNs \cite{Acharaya2018, Seal2021, Sharma2021}. More recently, state-of-the-art (SOTA) algorithms, such as U-Net, vision transformers, and graph neural networks, have been utilized to enhance accuracy even further \cite{Yang2023, Reazal2024, Luo2023}.

Despite recent progress, several challenges remain. The first is well-known: traditional deep learning models demand enormous amounts of data to achieve reliable performance \cite{smol_sample1}. In mental health (and healthcare more broadly), such high-quality annotated datasets are scarce \cite{biochallenge1, biochallenge2}. This scarcity stems from the labor-intensive process of data labeling, the limited number of patients with specific conditions, and the ethical concerns tied to patient privacy. The second challenge relates to the nature of EEG data. EEG signals are highly variable and prone to noise \cite{eeg_issue1, eeg_issue2}. Moreover, they are high-dimensional and complex, yet typically collected from relatively small samples, which reduces statistical power \cite{eeg_issue3}.   

To tackle these challenges, QML emerges as a promising direction. In theory, quantum models can capture complex, high-dimensional correlations and map data into richer Hilbert spaces \cite{hilber1,hilbert2}, which can help uncover hidden patterns inherent in brain waves. In addition, QML is being actively tested across different domains, showing advantages such as improved generalizability, noise robustness, and parameter efficiency over classical models \cite{Caro2022, GilFuster2024, Du2021}—qualities particularly relevant to healthcare. Owing to these benefits, several studies have already explored QML algorithms in healthcare \cite{masheshwari2022, Ullah2024}, with some focusing specifically on neuroinformatics \cite{Orka2025}. However, the potential of QML for depression diagnosis still remains unexplored.

With this gap in mind, we present a novel quanvolutional architecture, QuanvNeXt, in this study for EEG-based depression diagnosis. QuanvNeXt builds upon a previously developed 1D quantum convolution, or quanvolution, layer, which allows for modular usage by supporting arbitrary-channel input and a user-defined number of feature maps \cite{fqn}. The model incorporates a novel Cross Residual block inspired by ResNet \cite{resnet}, DenseNet \cite{densenet}, and ShuffleNet \cite{shufflenet}, which enables more efficient feature reuse, better gradient flow, and enhanced cross-channel interactions, ultimately improving both representational capacity and computational efficiency. QuanvNeXt achieved SOTA accuracy directly from time-domain EEG data without needing a time-frequency transformation. To validate the performance of QuanvNeXt, we provide a comparative analysis with other depression detection models from the literature, all of which were trained under similar constraints.

Beyond evaluating predictive performance, we examined the uncertainty and explainability of QuanvNeXt, as these aspects are critical for clinical reliability in depression diagnosis. To assess prediction uncertainty, we perturbed the input signals with small Gaussian noise and compared the model's stability against predictions on unaltered samples. This analysis enabled us to examine the confidence boundaries and assess robustness to input variability. For explainability, we employed explainable AI (XAI) frameworks that help reveal how the model processes EEG signals. Since QML models are inherently more complex and operate differently than classical deep learning models, demonstrating accuracy alone is insufficient. It is equally important to ensure that they identify relevant patterns and transparently distinguish between diagnostic classes. To this end, we applied two complementary post hoc XAI techniques: (i) hierarchical feature mapping to analyze the local contributions of each QuanvNeXt layer to the input data, and (ii) latent manifold projection to capture the global structure of learned representations. Together, these methods provide both fine-grained and holistic perspectives on the model's decision-making process.

The main contributions of this work are as follows:
\begin{itemize}
\item We conduct the first exploration of quantum models for mental-health analysis, motivated by prior evidence that quantum approaches can effectively model neural data.
\item We propose a novel Cross Residual Block designed specifically to enhance learning in time-series classification tasks.
\item We introduce an end-to-end quantum-based architecture that achieves SOTA performance on two depression datasets, without requiring any time–frequency transformation.
\end{itemize}

\section{Materials and Methods}
\subsection{Datasets}
We used two open-access datasets to evaluate QuanvNeXt. The first dataset (Dataset 1) was collected at the outpatient clinic of Hospital Universiti Sains Malaysia (HUSM) \cite{Mumtaz2016}. It comprises EEG recordings from 30 healthy controls (HCs) and 34 patients diagnosed with MDD. The EEG signals, sampled at 256 Hz, were captured using a 19-channel electro-gel sensor device, adhering to the international 10–20 system. The second dataset (Dataset 2) was collected at Lanzhou University Second Hospital, Gansu, China \cite{modma}. These data, sampled at 250 Hz, were acquired using a 128-electrode system following the 10–10 montage. Dataset 2 contains recordings from 25 MDD patients and 29 HCs.

\subsubsection{Cleaning}
For this study, we only worked with resting-state eyes closed (EC) data, as the absence of voluntary eye movements ensures a cleaner and more reliable representation of intrinsic brain activity \cite{barry2007}. However, both datasets still contained substantial noise, including various artifacts that could obscure brain waves and significantly hinder model learning. To mitigate these issues, we applied a data-cleaning process to both datasets prior to training, as briefed below.

Dataset 1 was missing six recordings (two HCs and four MDDs). The available data were preprocessed in EEGLAB \cite{delorme2004}, starting with a 0.5–70 Hz bandpass filter, followed by artifact subspace reconstruction (ASR) \cite{asr_new} to remove noisy segments and interpolate channels when necessary. A 50 Hz notch filter was applied after ASR, as it improved results in most cases. After automated cleaning, signals were manually inspected to trim residual artifacts. Some samples required channel interpolation even after manual trimming, but no more than two channels ($\le15\%$) were interpolated to maintain data integrity. Finally, signals were truncated to satisfy $\text{\textit{signal length}} = n \times \text{\textit{sampling rate}}$, where $n \in \mathbb{N}$. Excessively noisy samples or those requiring more than 15\% interpolation were discarded. After cleaning, 10 HC and 19 MDD recordings were retained.

In contrast, Dataset 2 was even noisier, requiring an updated cleaning pipeline. Similar to Dataset 1, we began with a 1–70 Hz bandpass filter. However, we could not directly use ASR after the bandpass filtering as all the readings had extreme 50 Hz noise from power line contamination that seeped well into the 47-53 Hz band, prompting ASR to remove whole signals. To resolve this, we applied two consecutive notch filters before ASR, primarily using IIR filters and switching to FIR only when passband ripples became excessive. After attenuating the power line noise, ASR was applied to remove bad segments and interpolate channels, with interpolation capped at 15\%. Signals were then manually inspected to reject residual noise, with further interpolation applied when necessary. Additional corrections for ocular and muscular artifacts were performed using blind source separation through the automatic artifact removal toolbox in EEGLAB. After cleaning, 10 MDD and 10 HC samples were retained.

\subsubsection{Preprocessing}
For both datasets, we segmented the cleaned data into overlapping 8-second windows to manage input lengths. The windowing function ensured signal continuity by adjusting for breaks and event boundaries. A 90\% overlap was applied to maximize samples. The resulting tensor shapes were (64, 19, 2056) for Dataset 1 and (64, 128, 2000) for Dataset 2, where each batch contained 64 windowed-EEG samples with 19 and 128 channels, respectively.

To avoid test-set contamination, we first applied a subject-wise 70–30 train–test split prior to windowing. Since windowing introduced class imbalance, we then applied random undersampling to match the majority class with the least frequent one. Finally, we introduced subject-independent, channel-wise Z-normalization to the entire dataset, using statistics from the train set only.

\subsection{Model Description}

\subsubsection{Quanv1D with Temperature Scaling}
\begin{figure}
    \centering
    \includegraphics[width=0.75\linewidth]{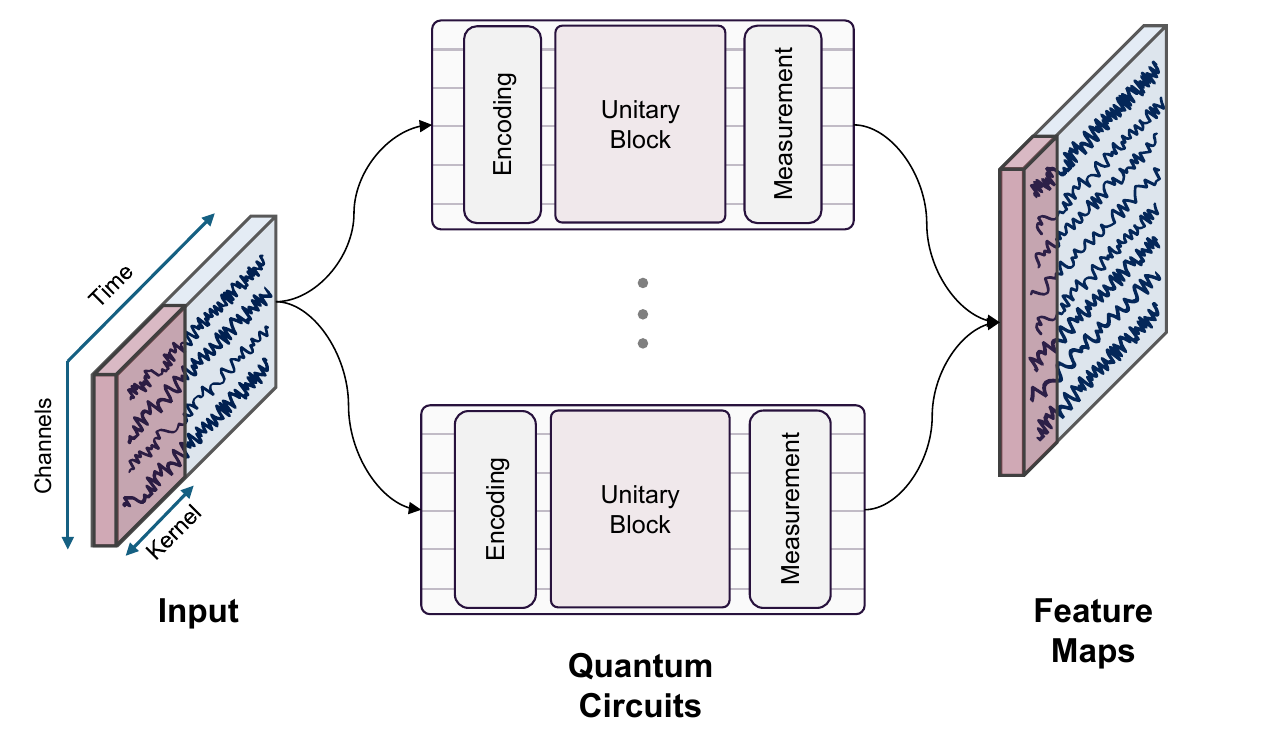}
    \caption{Workflow of the Quanv1D layer, which can handle arbitrary input and output sizes. After extracting patches, multiple trainable quantum circuits generate feature maps.}
    \label{fig:1dquanv}
\end{figure}

Our model is built entirely with the Quanv1D layer introduced by Orka \emph{et al.} \cite{fqn}. While Quanv1D functions similarly to the classical Conv1D layer, it replaces traditional filters or kernels with trainable quantum circuits (see Fig.~\ref{fig:1dquanv}). Unlike earlier quanvolutional layers, Quanv1D offers greater flexibility by supporting arbitrary-channel input, accommodating various kernel sizes, and enabling generating a user-defined number of feature maps, making it highly scalable.

Quanv1D layer is designed to emulate the behavior of a standard Conv1D layer, employing the same patching operations \cite{chellapilla2006high}. It processes inputs of shape $(N, C_{in}, L_{in})$, where $N$ denotes the batch size, $C_{in}$ the number of input channels (or dimensions), and $L_{in}$ the sequence length. After the quanvolution operation, the output takes the form $(N, C_{out}, L_{out})$, with $C_{out}$ specifying the number of output channels chosen by the user. The output length $L_{out}$ is determined by:

\begin{equation}
    L_{out} = \left\lfloor \frac{L_{in} + 2 \times p - d \times (k - 1) - 1}{s} + 1 \right\rfloor
    \label{hyper_eq}
\end{equation}

In this expression, $k$ represents the kernel size, $s$ the stride, $p$ the padding applied to both sides of the input, and $d$ the dilation factor, i.e., the spacing between kernel elements.

When applying QML to classical data, quantum circuits are generally structured into three main stages: encoding, manipulation, and measurement. The encoding stage maps classical information into a quantum feature space, preparing the data for subsequent quantum processing. We employed amplitude embedding, which encodes $2^n$ features into the amplitude vector of an $n$-qubit state \cite{amplitude1}, expressed as:

\begin{equation}
    | \psi \rangle = \sum_{i=1}^{2^n} \alpha_i |i \rangle
\end{equation}

In this equation, $\alpha_i$ denotes the components of the amplitude vector $\alpha$, while $|i\rangle$ corresponds to the computational basis states. For each quanvolutional filter, the input consists of $C_{in} \times k$ features, which requires $n = \left\lceil \log_2(C_{in} \times k) \right\rceil$ qubits for encoding. Before encoding, the features are normalized to ensure $|\alpha|^2=1$, and if needed, are padded with zeros to match the required dimension size. Orka \emph{et al.} applied $\sqrt{softmax(\alpha)}$ to ensure that the squared sum of all elements in the amplitude vector, $\alpha$, equals one \cite{fqn}. While this approach effectively promotes self-regularization by preventing extreme gradient values \cite{fqn}, it has a significant drawback: the fixed Softmax function does not allow control over the distribution of weights across different temporal features. As a result, it can sometimes produce an excessively smooth or overly sharp representation, which may hinder temporal feature extraction or eliminate critical variations necessary for classification. To address this limitation, we introduced temperature scaling, applying $\sqrt{softmax(\alpha/temp)}$ as part of the normalization process. The temperature parameter regulates the distribution of attention across time points, where higher values lead to more evenly distributed attention and lower values emphasize more discrete feature selection. This adjustment provides greater control over the balance between smooth and sharp transitions in time series data, ultimately improving feature representation.

After encoding, the quantum data is processed using a sequence of quantum gates in the second stage. The goal is to manipulate the quantum states in such a way as to find patterns and/or optimize solutions. The matrix representation of the unitary operator or gate chosen for this study is expressed as:

\begin{equation}
    U(\theta, \phi, \lambda) =
    \begin{pmatrix}
    \cos(\theta \frac{\pi}{2}) & -e^{i\lambda} \sin(\theta \frac{\pi}{2}) \\
    e^{i\phi} \sin(\theta \frac{\pi}{2}) & e^{i(\phi + \lambda)} \cos(\theta \frac{\pi}{2})
    \end{pmatrix}
\end{equation}

In this formulation, $\theta$ and $\lambda$ are treated as trainable parameters, whereas $\phi$ is fixed but initialized randomly. The reason is that, while $\phi$ plays a critical role in introducing phase shifts within the circuit, its gradient with respect to parameter updates is theoretically zero \cite{fqn}. For an $n$-qubit circuit, the unitary operator is applied individually to each qubit, creating a layer of unitaries. Repeating this layer $k$ times forms the final unitary block.

The final stage, decoding, measures the quantum states to recover classical outputs. Our decoding process is defined as follows:

\begin{equation}
    E_i = \langle \psi_o | Z_i | \psi_o \rangle
    \label{dec_1}
\end{equation}
\begin{equation}
    Z_i = I^{\otimes (i-1)} \otimes Z \otimes I^{\otimes (n-i)}
\end{equation}
\begin{equation}
    Z = \begin{pmatrix}
    1 & 0 \\
    0 & -1
    \end{pmatrix}
\end{equation}
\begin{equation}
    I = \begin{pmatrix}
    1 & 0 \\
    0 & 1
\end{pmatrix}
\end{equation}

Each qubit in the circuit is measured according to Eq. (\ref{dec_1}), where $E_i$ denotes the expectation value and $Z_i$ is the observable associated with the $i$-th qubit. Although the operations within each circuit or filter are performed in the complex domain, the resulting expectation values are real and constrained to the interval $[-1,1]$. Each expectation value is then assigned to a distinct output channel. Consequently, a filter with $n$ qubits produces $n$ feature maps. The total number of filters required is given by $\left\lfloor \frac{C_{\text{out}} + n - 1}{n} \right\rfloor$. If the number of generated feature maps exceeds the user-defined number of output channels, only the first $C_{out}$ maps are retained, with the remainder discarded.

\subsubsection{Cross Residual Block}
\label{sec:xRes}
\begin{figure}
    \centering
    \includegraphics[width=0.85\linewidth]{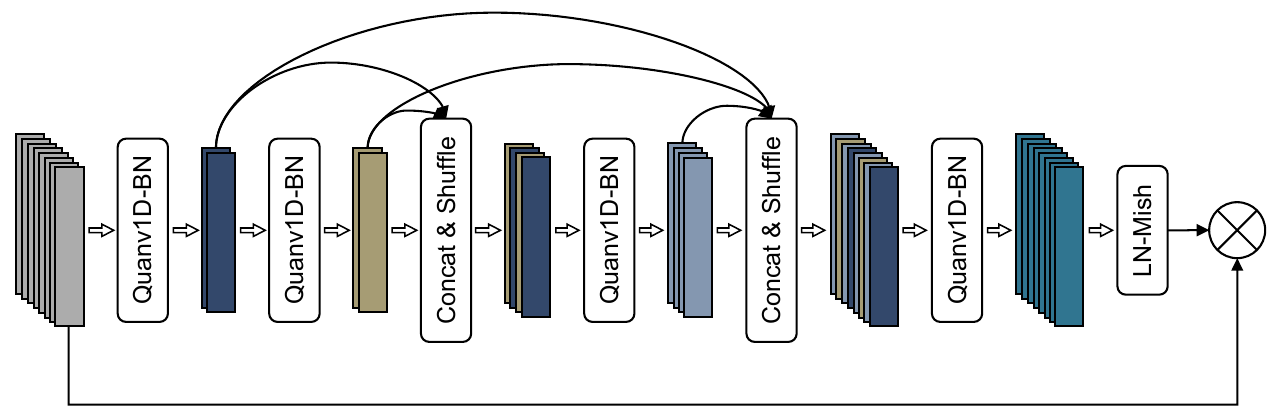}
    \caption{Cross Residual block. The block, proposed for time series, combines residual skip connections, dense feature aggregation, and channel shuffling for efficient representation learning. Note: BN: batch normalization; LN: layer normalization.}
    \label{fig:cross_residual}
\end{figure}

The Cross Residual block is the core unit of QuanvNeXt (see Fig. \ref{fig:cross_residual}), designed to combine the strengths of residual learning, dense feature aggregation, and channel shuffling. This hybrid design enables the model to capture both local dependencies (short-term trends) and global patterns (long-range dependencies) in time series data.

The residual skip connection, inspired by ResNet \cite{resnet}, enhances gradient flow stability and alleviates vanishing gradient issues. This mechanism not only stabilizes optimization but also ensures that the model retains raw/original signal information, improving robustness against distortions. Meanwhile, feature aggregation and concatenation, motivated by DenseNet \cite{densenet}, promote feature reuse by feeding multi-scale representations forward through the block. This leads to a richer hierarchical representation of temporal dynamics, where both fine-grained fluctuations and broader trends are captured. Another key advantage of feature reuse is that it allows the network to generate a larger set of feature maps without increasing the number of learnable parameters.

To preserve feature diversity, we employed channel shuffling inspired by ShuffleNet \cite{shufflenet}. In deeper models, certain filters often become biased toward specific input channels, effectively learning only a subset of the information rather than the broader context. Channel shuffling enhances contextual awareness and prevents channel-feature isolation by mixing cross-channel features. As a result, the layers are encouraged to model the full temporal input, rather than being restricted to narrow patterns. While similar diversity could be achieved by widening Quanv1D layers, channel shuffling provides a non-trivial alternative that improves representational capacity without increasing the number of parameters. Our block incorporates two channel shuffling layers. The first operates with four groups, while the second operates with eight. This progressive shuffling facilitates better cross-channel information exchange at different granularities.

Finally, for training stability across varying sequence lengths, we employed layer normalization instead of batch normalization, as the latter depends on batch-level statistics and can become unstable with small batch sizes \cite{layernorm}. Additionally, we adopt the Mish activation function in place of ReLU, since Mish preserves small negative values and yields smoother gradients, enabling more effective weight updates \cite{mish}.

\subsubsection{QuanvNeXt}
The overall architecture is outlined in Table~\ref{tab:model_archi}. Note that each component or layer used in constructing the blocks is implemented with Quanv1D layers. The first component is the Windowed Embedding layer, which functions as a learnable feature extractor by transforming raw time series data into structured feature representations of user-defined dimensions. Unlike naive downsampling, which risks losing critical information, this embedding step preserves necessary information within each non-overlapping segment and projects high-dimensional data into a compact latent space.

\begin{table}
\centering
\caption{The architecture of QuanvNeXt with its hyperparameters.}
\label{tab:model_archi}
\resizebox{0.7\columnwidth}{!}{%
\begin{tabular}{@{}llll@{}}
\toprule
 & \textbf{Block} & \textbf{Hyperparameters} & \textbf{Output shape} \\ \midrule 
 & Windowed Embedding & $k=8$, $s=8$ & (32, 256) \\ \cmidrule(l){2-4} 
\begin{tabular}{c} \rotatebox{90}{Dataset 1} \end{tabular} & \begin{tabular}[c]{@{}l@{}}Cross Residual Blocks \\ ($\times 4$)\end{tabular} & \begin{tabular}[c]{@{}l@{}}$k=7$, $s=1$, $p=3$, $temp=1.5$\\ $k=17$, $s=1$, $p=8$, $temp=1.2$\\ $k=11$, $s=1$, $p=5$, $temp=0.8$\\ $k=7$, $s=1$, $p=3$, $temp=0.5$\end{tabular} & (32, 256) \\ \cmidrule(l){2-4} 
 & Windowed Projection & $k=8$, $s=8$ & (2, 32) \\ \midrule

 & Windowed Embedding & $k=8$, $s=8$ & (8, 250) \\ \cmidrule(l){2-4} 
\begin{tabular}{c} \rotatebox{90}{Dataset 2} \end{tabular} & \begin{tabular}[c]{@{}l@{}}Cross Residual Blocks \\ ($\times 4$)\end{tabular} & \begin{tabular}[c]{@{}l@{}}$k=7$, $s=1$, $p=3$, $temp=1.5$\\ $k=15$, $s=1$, $p=7$, $temp=1.2$\\ $k=9$, $s=1$, $p=4$, $temp=0.8$\\ $k=7$, $s=1$, $p=3$, $temp=0.5$\end{tabular} & (8, 250) \\ \cmidrule(l){2-4} 
 & Windowed Projection & $k=8$, $s=8$ & (2, 31) \\ \bottomrule
\multicolumn{4}{l}{\small $k$: kernel; $s$: stride; $p$: padding; $d$: dilation; $temp$: temperature scaling}
\end{tabular}%
}
\end{table}

The second component of our architecture is the proposed Cross Residual block. Through extensive trial-and-error runs, we observed that stacking four such blocks yields significant performance improvements. In addition to utilizing the inherent parameter efficiency and self-regularization properties of Quanv1D \cite{fqn}, our Cross Residual block facilitates effective temporal feature learning without a substantial increase in parameter count. Furthermore, we incorporate temperature scaling within these blocks to modulate feature emphasis across layers (see Table~\ref{tab:model_archi}). Specifically, the temperature is initialized at a higher value to capture broad contextual dependencies and is gradually reduced to refine local feature representations progressively. This strategy ensures a smooth transition from global to fine-grained temporal patterns. Overall, the introduction of this novel block sets QuanvNeXt apart from other SOTA depression classifiers, as will be discussed in later sections.

Finally, the Windowed Projection module maps the learned feature representations from preceding layers into a class-specific latent space. While similar in design to the embedding block, this module projects features into a class-relevant dimension. The goal of this component is to learn distinct feature representations for each class, thereby improving the overall discriminative power and classification performance of the model. For the final classification, we applied global average pooling over the sequence length dimension.

\subsection{Evaluation Metrics}
\subsubsection{MCC} 
Matthews Correlation Coefficient (MCC) is a balanced measure that accounts for all four confusion matrix components, producing values in the range of $-1$ (total misclassification) to $1$ (perfect classification).
    \begin{equation}
        \begin{gathered}
        MCC = (TP \times TN) - (FP \times FN) /\\
        \sqrt{(TP + FP)(TP + FN)(TN + FP)(TN + FN)}
        \end{gathered}
    \end{equation}

Here, TP, TN, FP, and FN denote true positives, true negatives, false positives, and false negatives, respectively.

\subsubsection{AUC-ROC} 
Area Under the Receiver Operating Characteristic Curve (AUC-ROC) evaluates a classifier's ability to distinguish between positive and negative cases by plotting the True Positive Rate (TPR) against the False Positive Rate (FPR) across various decision thresholds. AUC-ROC values range from 0.5 (random guessing) to 1 (perfect classifier).
    \begin{equation}
        AUC = \int_{0}^{1} TPR(FPR) \ d(FPR)
    \end{equation}    
    \begin{equation}
        FPR = \frac{FP}{FP + TN}
    \end{equation}    
    \begin{equation}
        TPR = \frac{TP}{TP + FN}
    \end{equation}

\subsubsection{ECE} 
Expected Calibration Error (ECE) measures the expected difference between a model's confidence and actual accuracy by partitioning predictions into bins and computing a weighted average of the absolute calibration error in each bin. Lower ECE values indicate better calibration, with 0 representing perfect calibration.
    \begin{equation}
        \text{ECE} = \sum_{i=1}^{M} \frac{|B_i|}{n} \left| acc(B_i) - conf(B_i) \right|
    \end{equation}
    Here, $M$ is the number of bins, $|B_i|$ is the number of predictions in bin $i$, $n$ is the total number of predictions, and $acc(B_i)$ and $conf(B_i)$ represent the accuracy and average confidence in bin $i$, respectively.

\section{Results and Discussion}
\subsection{Classification Performance}
In this section, we evaluated the classification performance of our proposed model, QuanvNeXt. Additionally, we compared it to other SOTA depression detection models that analyze raw EEG signals rather than those relying on image transformations like scalograms or spectrograms. The metrics used in the comparison are detailed in the Appendix.

Although no other quantum models were explored for depression detection prior to our study, we decided to adopt a few for completeness in the comparison. We identified three convolution-based quantum models—FQN \cite{fqn}, QuanvNet \cite{quanvnet}, and ESQN \cite{esqn}—that operate directly on time-series data without requiring visual transformation. However, among these, QuanvNet and ESQN cannot process multichannel or multivariate time series. As such, only FQN was retained for our comparative analysis.

Using PyTorch, we ran our experiments on an NVIDIA GeForce RTX 4060 Ti 16GB GPU. We implemented the compared models from scratch, following the methodologies described in their respective papers. We trained each model for 300 epochs with a batch size of 64 and optimized them with NAdam \cite{nadam}. Besides the learning rate (0.00015 for Dataset 1 and 0.0025 for Dataset 2), we relied on PyTorch’s default settings. We saved the model after every epoch instead of using a dedicated scheduler. Tables \ref{tab:comp_anal1} and \ref{tab:comp_anal2} list the models that performed the best on the test sets.

\begin{table}
\centering
\caption{Performance evaluation on the test set of Dataset 1.}
\label{tab:comp_anal1}
\resizebox{0.65\textwidth}{!}{%
\begin{tabular}{@{}lllll@{}}
\toprule
\textbf{Model} & \textbf{Parameters} & \textbf{Accuracy} & \textbf{AUC-ROC} & \textbf{MCC} \\ \midrule
1DCNN-Transformer \cite{1dcnn-transformer} & 37946 & 0.6288 & 0.6558 & 0.2627 \\
xCNN \cite{xcnn} & 1619 & 0.7321 & 0.7922 & 0.4646 \\
1DCNN-LSTM \cite{1dcnn-lstm} & 413642 & 0.7127 & 0.7555 & 0.4838 \\
STCNN \cite{stcnn} & 454626 & 0.7618 & 0.8052 & 0.532 \\
1DCNN-GRU \cite{1dcnn-gru} & 30626 & 0.7679 & 0.8743 & 0.5449 \\
TanhReLU-CNN \cite{tanhrelu} & 6233704 & 0.7863 & 0.8058 & 0.5731 \\
Ex-1DCNN \cite{ex-1dcnn} & 622 & 0.7883 & 0.8215 & 0.5774 \\
MADNet \cite{madnet} & 88936 & 0.8262 & 0.9078 & 0.6595 \\
2DCNN-LSTM \cite{2dcnn-lstm} & 217090 & 0.8681 & 0.914 & 0.7385 \\
DSNet \cite{dsnet} & 11106 & 0.9192 & 0.9935 & 0.8496 \\
FQN \cite{fqn} & 1994 & 0.9254 & 0.9254 & 0.8507 \\
InceptionTime \cite{inceptiontime} & 1700989 & 0.956 & 0.9841 & 0.9141 \\
QuanvNeXt (proposed) & 6144 & 0.9591 & 0.9931 & 0.9197 \\ 
\bottomrule
\end{tabular}%
}
\end{table}

\begin{table}
\centering
\caption{Performance evaluation on the test set of Dataset 2.}
\label{tab:comp_anal2}
\resizebox{0.65\textwidth}{!}{%
\begin{tabular}{@{}lllll@{}}
\toprule
\textbf{Model} & \textbf{Parameters} & \textbf{Accuracy} & \textbf{AUC-ROC} & \textbf{MCC} \\ \midrule
TanhReLU-CNN \cite{tanhrelu} & 6288204 & 0.5 & 0.5 & 0 \\
1DCNN-Transformer \cite{1dcnn-transformer} & 48410 & 0.5854 & 0.5431 & 0.1786 \\
2DCNN-LSTM \cite{2dcnn-lstm} & 217090 & 0.6098 & 0.4717 & 0.2707 \\
Ex-1DCNN \cite{ex-1dcnn} & 2257 & 0.5854 & 0.3183 & 0.3055 \\
DSNet \cite{dsnet} & 17890 & 0.5854 & 0.8037 & 0.3055 \\
MADNet \cite{madnet} & 158696 & 0.6341 & 0.5806 & 0.3073 \\
1DCNN-GRU \cite{1dcnn-gru} & 37602 & 0.5976 & 0.5146 & 0.3288 \\
1DCNN-LSTM \cite{1dcnn-lstm} & 442378 & 0.6829 & 0.699 & 0.3892 \\
xCNN \cite{xcnn} & 1579 & 0.7195 & 0.7597 & 0.4423 \\
STCNN \cite{stcnn} & 479042 & 0.7317 & 0.7192 & 0.4685 \\
FQN \cite{fqn} & 842 & 0.8415 & 0.8415 & 0.6833 \\
InceptionTime \cite{inceptiontime} & 1742082 & 0.878 & 0.9328 & 0.7597 \\
QuanvNeXt (proposed) & 2488 & 0.8902 & 0.95 & 0.7826 \\ 
\bottomrule
\end{tabular}%
}
\end{table}

QuanvNeXt stood out for its strong performance across all evaluation metrics while maintaining exceptional parameter efficiency. On Dataset 1, it achieved the highest accuracy (0.9591) and MCC (0.9197), slightly outperforming InceptionTime, which required over 277$\times$ more parameters. Similarly, in Dataset 2, QuanvNeXt delivered the best overall results, with an accuracy of 0.8902, a MCC of 0.7826, and an AUC-ROC of 0.95, while using only 2488 parameters. Models like DSNet and FQN also demonstrated competitive results with low parameter counts, but QuanvNeXt consistently outperformed them. Overall, QuanvNeXt showed the most balanced performance despite its lightweight nature.

\subsection{Ablation Study}
Despite FQN and QuanvNeXt having similar “fully quanvolutional” architectures, QuanvNeXt gained an edge thanks to our proposed Cross Residual block. To quantify its contribution, we conducted an ablation study to evaluate how much this block influenced overall performance.

As shown in Table~\ref{tab:ablation}, each individual module within the block contributes substantially to overall performance, collectively improving temporal representation learning. This observation aligns with the theoretical motivations discussed in Section~\ref{sec:xRes}. Notably, the seemingly simple channel shuffling module has the largest impact, causing a drop in accuracy of over 8\% on both datasets when removed. The underlying reason can be explained as follows.

\begin{table}
\centering
\caption{Ablation study across datasets.}
\label{tab:ablation}
\resizebox{0.45\columnwidth}{!}{%
\begin{tabular}{@{}lll@{}}
\toprule
 & \textbf{Task} & \textbf{Performance} \\ 
 & & \textbf{decrease} \\ \midrule
\multirow{3}{*}{\begin{tabular}{c} \rotatebox{90}{Dataset 1} \end{tabular}} & Without skip connection & 6.54\% \\ \cmidrule(l){2-3}
 & Without feature aggregation & 1.74\% \\ \cmidrule(l){2-3}
 & Without channel shuffling & 8.69\% \\ \midrule
\multirow{3}{*}{\begin{tabular}{c} \rotatebox{90}{Dataset 2} \end{tabular}} & Without skip connection & 6.09\% \\ \cmidrule(l){2-3}
 & Without feature aggregation & 3.65\% \\ \cmidrule(l){2-3}
 & Without channel shuffling & 9.75\% \\ \bottomrule 
\end{tabular}%
}
\end{table}

Quanv1D utilizes all available wires in a circuit or filter to produce multiple maps, establishing a one-to-one correspondence between each wire and a feature map. While this design is highly parameter-efficient (a finding supported by both the original study \cite{fqn} and our experiments), it comes at the cost of reduced feature diversity. Since multiple maps originate from the same source or filter, the resulting features, although varied, are inherently homogeneous. Channel shuffling addresses this limitation by mixing features across different layers, preventing over-reliance on specific features. This mechanism effectively restores diversity and enhances the representational capacity of the model.

\subsection{Post Hoc Explainability}
In this paper, we implemented two XAI approaches: hierarchical feature mapping and latent manifold projection.

\begin{figure}
    \centering
    \includegraphics[width=\linewidth]{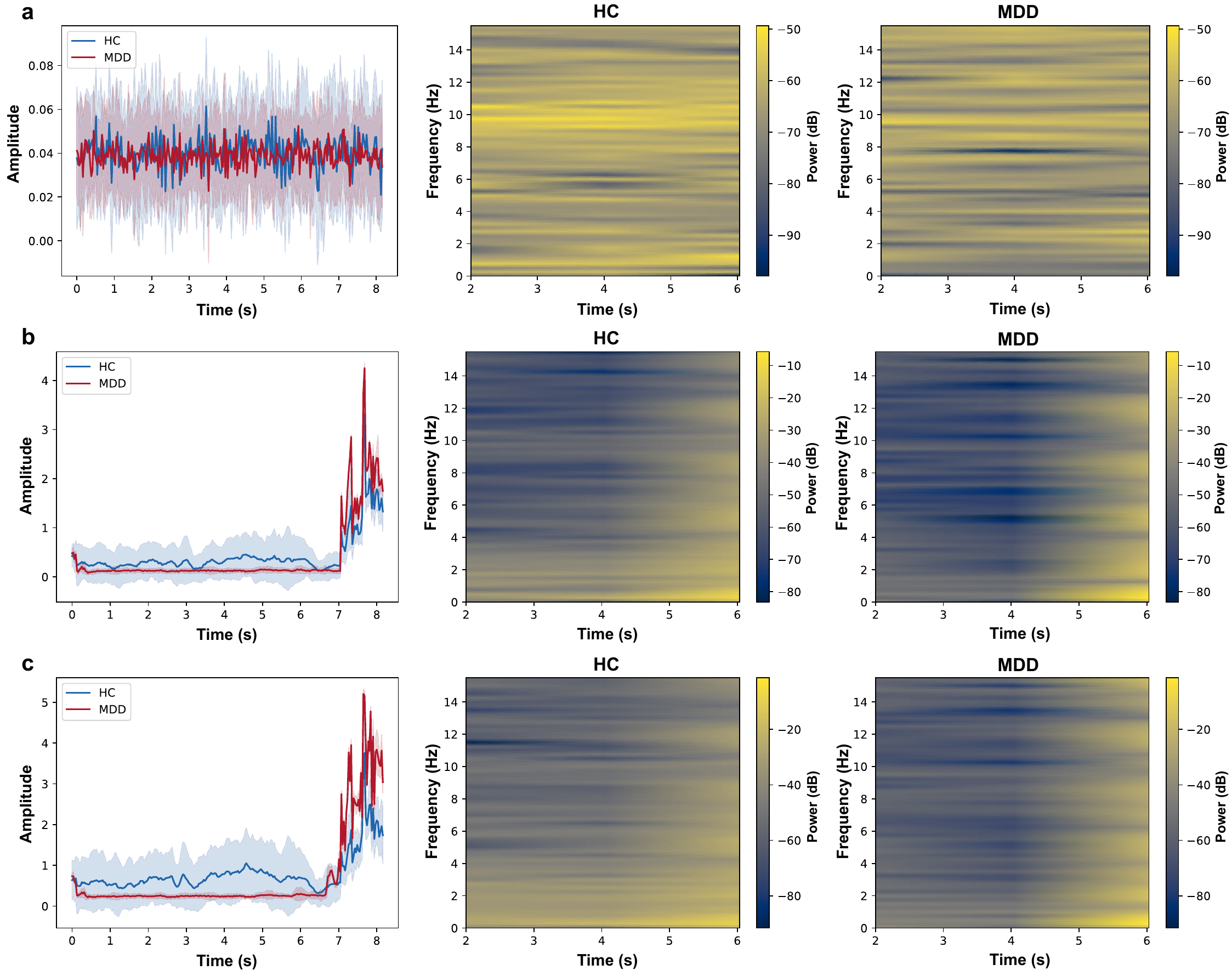}
    \caption{Hierarchical feature maps of HC and MDD subjects in Dataset 1, showing time-domain representations and corresponding spectrograms of the embedded activations. (a) Features after the Embedding layer; (b) Features in the middle of the network, i.e., after the 2nd Cross Residual block; (c) Features towards the end of the network, i.e., after the 4th and final Cross Residual block. Note: time and frequency axes correspond to latent representations derived from the embedding process, not the raw EEG domain.}
    \label{fig:feature_map1}
\end{figure}

\begin{figure}
    \centering
    \includegraphics[width=\linewidth]{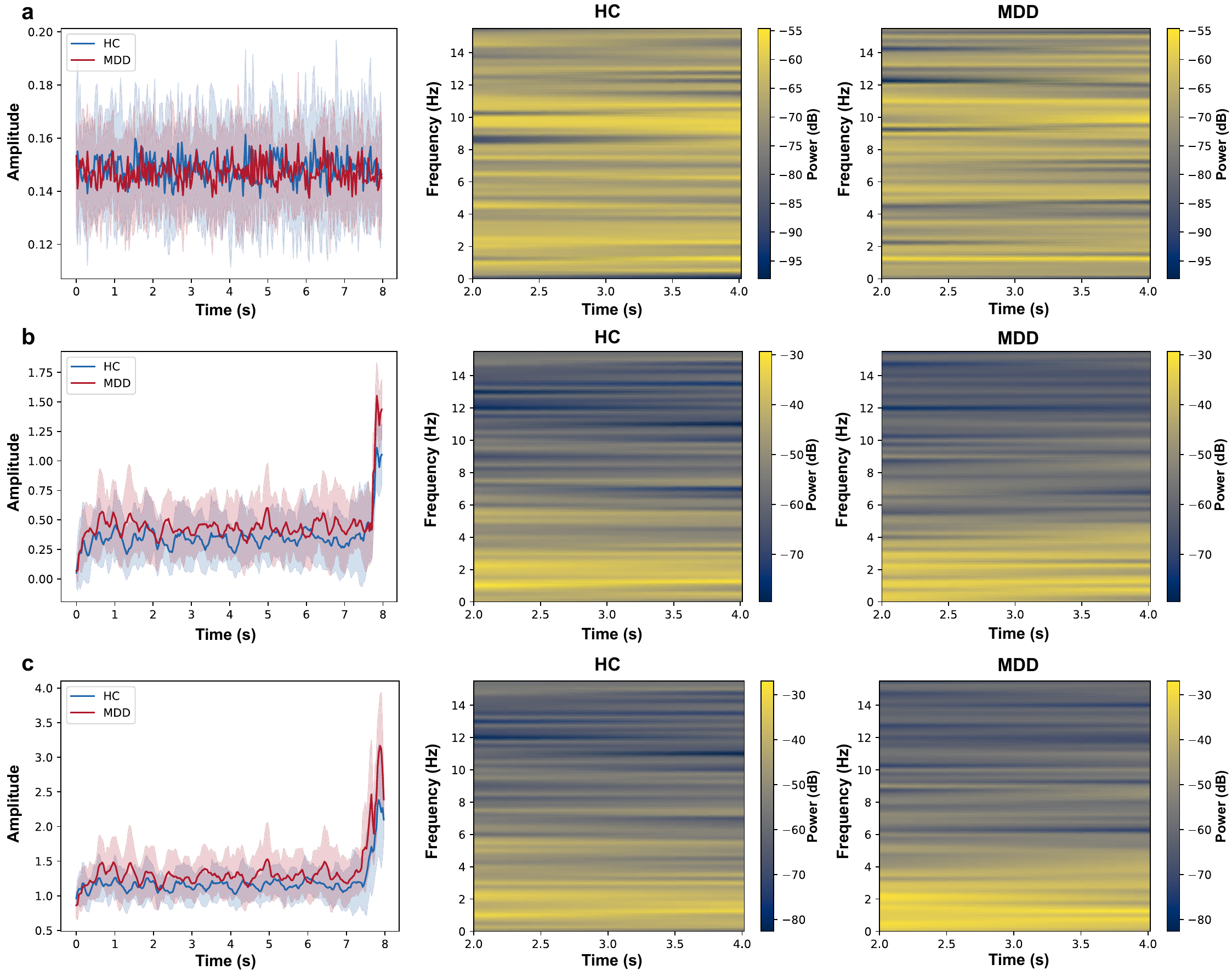}
    \caption{Hierarchical feature maps of HC and MDD subjects in Dataset 2, showing time-domain representations and corresponding spectrograms of the embedded activations. (a) Features after the Embedding layer; (b) Features in the middle of the network, i.e., after the 2nd Cross Residual block; (c) Features towards the end of the network, i.e., after the 4th and final Cross Residual block. Note: time and frequency axes correspond to latent representations derived from the embedding process, not the raw EEG domain.}
    \label{fig:feature_map2}
\end{figure}

To understand the internal representations learned by the model, we employed hierarchical feature mapping, where we extracted feature activations from different layers or blocks of the network and visualized their temporal and spectral characteristics (see Figs. \ref{fig:feature_map1} and \ref{fig:feature_map2}). The goal was to analyze how the model transforms raw input into meaningful representations progressively.

Figures \ref{fig:feature_map1} and \ref{fig:feature_map2} illustrate the temporal and spectral characteristics of internal representations learned by the QuanvNeXt model at three different processing stages. To provide an interpretable view, we selected one representative feature channel per block based on the highest variance across samples. For each class (HC vs. MDD), 16 random samples were drawn from the test sets, and their activations were averaged in time. The solid lines represent the mean activation across these samples, while the shaded regions denote the corresponding standard deviation. For both figures, the left subpanel shows the temporal feature maps, while the middle and right subpanels show the spectral decompositions or spectrograms of the mean activations for the two classes. It is important to note that the horizontal axis labeled as ``time" and the vertical axis labeled as ``frequency" in these plots do not correspond to the raw EEG domain. Here, the time axis reflects progression along the embedded sequence, and the frequency axis captures oscillation rates of the embedded activations, rather than true EEG frequency bands. As such, readers should interpret these axes as latent spectrotemporal dynamics of learned features, not literal seconds or Hertz.

Across both datasets, a consistent pattern is that at shallower levels, the activations show weaker separability between classes, with overlapping temporal profiles and spectrograms. However, as features propagate deeper into the model, the differences between HC and MDD representations become more distinct, reflecting the ability of QuanvNeXt to extract discriminative patterns. On Dataset 1, the differentiating factor appears to be stronger temporal activations in HC, with higher-amplitude fluctuations visible in the time-domain representations. This is reflected in the corresponding spectrograms, where HC activations show dispersed power across the latent spectral range rather than being confined to specific regions (see Fig. \ref{fig:feature_map1}c). In contrast, MDD representations display more concentrated activity in the lowest latent frequencies (0–1 Hz), making these regions more prominent (see Fig. \ref{fig:feature_map1}c). On Dataset 2, a somewhat different pattern emerges. Both HC and MDD show temporal activations of comparable strength, but the MDD class exhibits greater power in the lower latent frequency range (0–3 Hz) compared to HC (see Fig. \ref{fig:feature_map2}c). Although the representations look similar, QuanvNeXt still finds subtle differences that allow it to separate the classes.

\begin{figure}
    \centering
    \includegraphics[width=\linewidth]{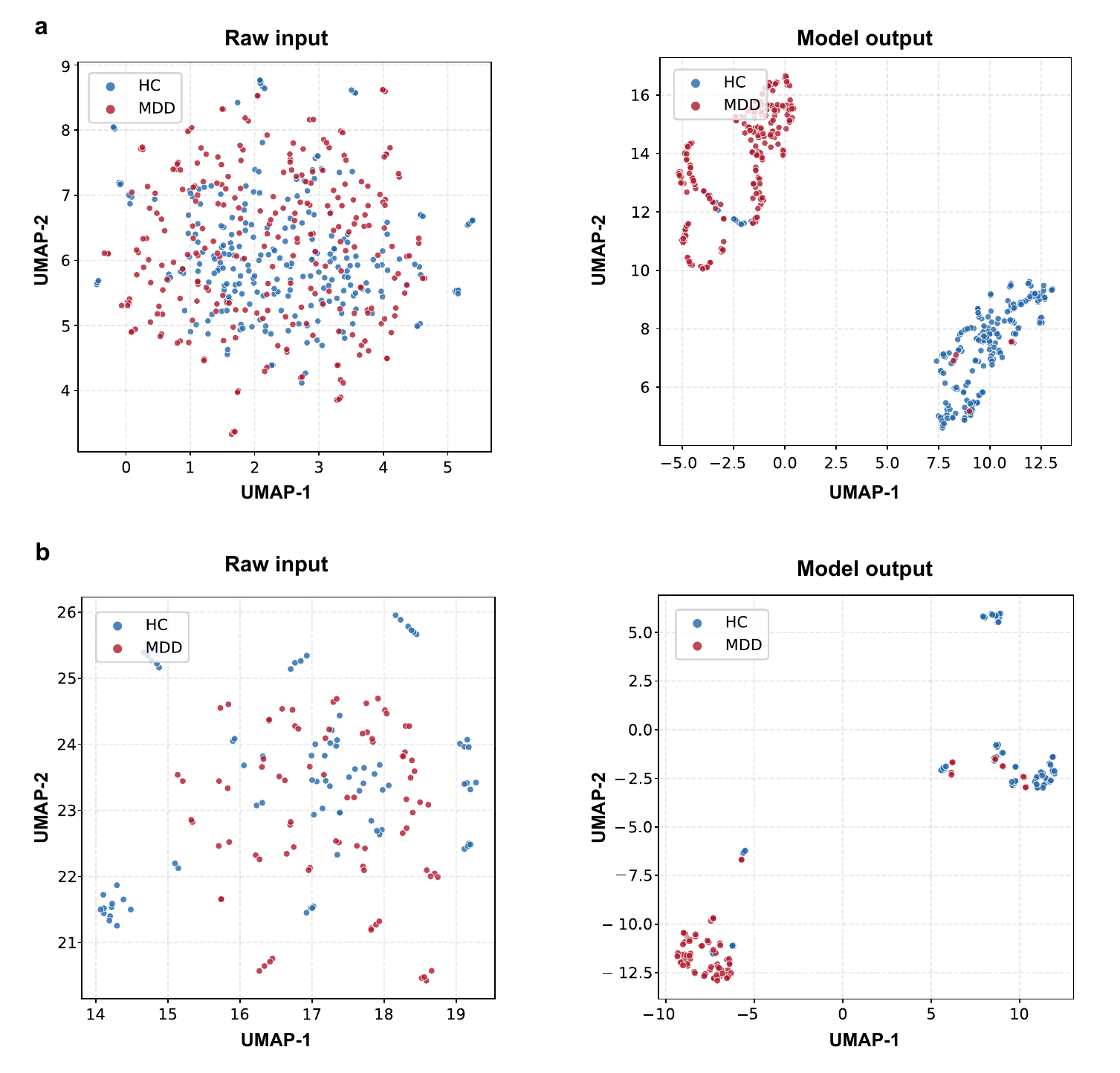}
    \caption{Comparison of raw EEG features and model-learned latent representations using UMAP. Points represent individual subjects, colored by group (HC vs MDD). The left column shows raw input data, and the right column shows the features after model projection. (a) Dataset 1; (b) Dataset 2.}
    \label{fig:latent_map}
\end{figure}

In addition to visualizing feature maps, we sought to understand how the model structures data within its learned representation space (see Fig. \ref{fig:latent_map}). To achieve this, we employed latent manifold projection, where we projected both raw input features and deep latent representations onto a low-dimensional space using Uniform Manifold Approximation and Projection (UMAP) \cite{umap}.

We first applied UMAP to the raw input EEG signals from the test sets, providing a baseline understanding of how separable the two classes are before any learning occurs. Next, we extracted deep feature representations after the Windowed Projection layer (which produces class-specific latent embeddings) and applied UMAP again. By comparing these two projections, we aimed to assess how well the model differentiates between classes through learned manifolds. Figure~\ref{fig:latent_map} demonstrates that while the raw EEG data shows no clear separation pattern between HC subjects and those with depression, the projection layer organizes the data in a way that reveals this distinction much more clearly. The well-separated clusters in the right-hand plots indicate that the model has learned distinctive EEG patterns associated with each mental state. It should be noted that, as QuanvNeXt is not perfectly accurate, a small number of samples are misclassified.

\subsection{Uncertainty Quantification}
To quantify how the model responds to minor variations in EEG data, we introduced small perturbations to input samples and observed the stability of predictions (see Table~\ref{tab:uncertainty_summary}). Given an input sample, we generated $n=50$ perturbed versions by adding small Gaussian noise: $x^\prime = x + \epsilon \cdot \mathcal{N} (0, 1)$, where $\epsilon \in \{0.1,0.05,0.01\}$ controls the magnitude of the noise. The mean prediction across perturbations was taken as the final class estimate, while uncertainty was measured using the variability of predictions. Model calibration was assessed using ECE, where $n_{bins}=10$. This setup allowed us to evaluate prediction stability and the alignment between model confidence and actual performance.

\begin{table}
\centering
\caption{Summary of uncertainty analysis on the test sets, computed for both correct and incorrect predictions.}
\label{tab:uncertainty_summary}
\resizebox{0.8\textwidth}{!}{%
\begin{tabular}{llllll}
\toprule
 & $\mathbf{\epsilon}$ & \textbf{Accuracy} & \textbf{Mean uncertainty} & \textbf{Mean uncertainty} & \textbf{ECE} \\
 &  & \textbf{(95\% CI)} & \textbf{(Correct)} & \textbf{(Incorrect)} &  \\
\midrule
\multirow{3}{*}{\begin{tabular}{c} \rotatebox{90}{Dataset 1} \end{tabular}}
 & 0.1 & 0.9652 (0.9518--0.9750) & 0.0181 & 0.0963 & 0.0436 \\ \cmidrule(l){2-6}
 & 0.05 & 0.9601 (0.9460--0.9707) & 0.0094 & 0.0557 & 0.0393 \\ \cmidrule(l){2-6}
 & 0.01 & 0.9591 (0.9448--0.9698) & 0.0021 & 0.0116 & 0.0373 \\
\midrule
\multirow{3}{*}{\begin{tabular}{c} \rotatebox{90}{Dataset 2} \end{tabular}} 
 & 0.1  & 0.8659 (0.7755--0.9234) & 0.0063 & 0.0311 & 0.1159 \\ \cmidrule(l){2-6}
 & 0.05  & 0.8780 (0.7899--0.9324) & 0.0047 & 0.0151 & 0.1025 \\ \cmidrule(l){2-6}
 & 0.01  & 0.8902 (0.8044--0.9412) & 0.0011 & 0.0023 & 0.1413 \\
\bottomrule
\multicolumn{6}{l}{\small $\epsilon$: magnitude of perturbation; CI: confidence interval}
\end{tabular}%
}
\end{table}

The results (see Table~\ref{tab:uncertainty_summary}) show that QuanvNeXt exhibits high robustness on Dataset 1, as accuracy stayed above 95\% even after adding noise. The low ECE values indicate that predicted confidences closely match actual accuracy, meaning the model is well-calibrated. Moreover, mean uncertainty is consistently higher for incorrect predictions than correct ones, suggesting that the uncertainty measure effectively identifies less reliable outputs. Additionally, Table~\ref{tab:uncertainty_summary} showcases that accuracy slightly increases with higher-magnitude noise, likely due to a small smoothing effect. Sometimes, tiny amounts of noise can help the model avoid overconfident mistakes, similar to test-time data augmentation \cite{tta}. This finding also aligns with prior studies on QML \cite{Du2021, robust2} and classical neural networks \cite{controlled_noise1}, which suggest that noise can improve data learning. In contrast, the results of Dataset 2 are less consistent. Although accuracy remains relatively high, ECE fluctuated. This likely reflects the small test size of Dataset 2, which makes calibration estimates less stable. With fewer samples, even small changes in predictions can cause large fluctuations in ECE, which may not reflect true model behavior. However, despite these inconsistencies, QuanvNeXt is still reliable, as mean uncertainties for correct predictions are consistently lower than for incorrect ones, similar to Dataset 1. Overall, the results suggest that uncertainty estimates are more stable and reliable on larger test sets (Dataset 1), whereas conclusions on smaller cohorts (Dataset 2) should be interpreted with caution.

\section{Conclusion and Future Work}
For this study, our objective was to propose a new quanvolutional neural network that is not only efficient in terms of accuracy and parameter usage but also explainable and reliable. The proposed model, QuanvNeXt, achieved SOTA accuracy and MCC on both evaluated datasets, owing to the novel Cross Residual block. This performance was achieved despite QuanvNeXt being, on average, 500$\times$ lighter than the second-best model in the comparative analysis. Post hoc XAI analysis confirmed that QuanvNeXt effectively captures class-differentiating patterns by focusing on relevant temporal and frequency characteristics. Furthermore, uncertainty quantification revealed that QuanvNeXt exhibits a degree of self-awareness in its predictions. The significant differences in uncertainty levels between correct and incorrect predictions suggest that the model can reliably assess its confidence, effectively `knowing when it knows.'

Moving forward, we identified two promising directions for future research. The first avenue concerns quantum hardware. QuanvNeXt is designed as a quantum-classical hybrid neural network, simulated in a classical computer. While most learnable layers are quantum, classical normalization and activation layers provide the necessary nonlinearity in the otherwise linear quantum domain. As such, our model is not fully quantum. Moreover, we employed analytical implementations of amplitude embedding and expectation value calculation in this study. Although these choices reduced circuit depth and computational overhead, they are not directly realizable on actual quantum devices. Thus, our next step is to evaluate QuanvNeXt on real quantum hardware, with suitable adjustments in encoding and measurement strategies, to validate its practical feasibility.

The second direction involves enhancing the reliability and trustworthiness of QuanvNeXt. While we explored some basic XAI methods, more advanced approaches are needed. In future work, we plan to incorporate advanced post hoc techniques such as Grad-CAM \cite{gradcam}, LRP \cite{lrp}, and SHAP \cite{shap}. Additionally, we could not quantify uncertainty accurately due to the limited sample size of Dataset 2. To address this, we intend to evaluate the model on larger cohorts of individuals with and without depression to understand its reliability and accuracy better. However, given the scarcity of large, high-quality annotated mental health datasets, we also aim to explore self-supervised \cite{selfsuper} and few-shot \cite{fewshot} learning approaches to assess whether they can further improve QuanvNeXt's generalizability in low-resource clinical settings.

\section*{Data Availability}
Both open-access datasets used in this study are available on Figshare (\url{https://doi.org/10.6084/m9.figshare.4244171.v2}) and ReShare (\url{https://dx.doi.org/10.5255/UKDA-SN-854301}), respectively.

\section*{Author Contributions}
\textbf{Nabil Anan Orka}: Conceptualization, Formal analysis, Investigation, Methodology, Software, Visualization, Writing -- original draft. \textbf{Ehtashamul Haque}: Conceptualization, Investigation, Data curation, Writing -- original draft. \textbf{Maftahul Jannat}: Validation, Resources, Writing -- Review \& Editing. \textbf{Md Abdul Awal}: Supervision, Writing -- Review \& Editing. \textbf{Mohammad Ali Moni}: Validation, Writing -- Review \& Editing, Supervision, Project administration.
\bibliographystyle{elsarticle-num} 
\bibliography{ref}






\end{document}